\documentclass[letterpaper, 10 pt, conference]{ieeeconf}  

\IEEEoverridecommandlockouts                              
\usepackage{siunitx}
\usepackage[table,x11names]{xcolor}
\usepackage[table,xcdraw]{xcolor}
\usepackage[dvipsnames]{xcolor}
\usepackage{svg}
\usepackage{graphicx}
 \graphicspath{{./figures/}}

\usepackage{amsmath}
\usepackage{tcolorbox}
\usepackage{enumerate}
\let\labelindent\relax
\usepackage[shortlabels]{enumitem}\usepackage{amssymb}
\usepackage{latexsym}
\usepackage{url}
\usepackage{cite}
\usepackage{relsize}
\usepackage{multirow}
\usepackage{afterpage}
\usepackage{ifthen}
\usepackage{graphicx}
\usepackage{algpseudocode,algorithm,algorithmicx}
\usepackage{subfigure}
\usepackage{flushend}
\usepackage{epstopdf}
\usepackage{stackengine}
\usepackage{textcomp} 
\usepackage{pifont}
\usepackage{threeparttable,booktabs}
\usepackage{gensymb}
\usepackage{booktabs}
\usepackage{makecell}
\usepackage{hhline}
\usepackage{courier}
\usepackage{lipsum}
\usepackage{diagbox}
\usepackage{xcolor} 
\usepackage{xspace}
\usepackage{wrapfig}

\usepackage{placeins}

\usepackage{array}
\newcolumntype{C}{>{\centering\arraybackslash}p{0.9cm}}

\newcommand{\etal}{\textit{et al.}}
\newcommand{\datasetName}{RobotDesign1M}

\makeatletter
\let\NAT@parse\undefined
\makeatother
\usepackage[bookmarks=false,linkcolor=blue, urlcolor=blue, citecolor=blue]{hyperref} 

\hypersetup{
    colorlinks=true,
    linkcolor=red,
    filecolor=magenta,      
    urlcolor=blue,
    pdfstartview={FitH},
    citecolor =blue
    }

\title{\LARGE \bf RobotDesign1M: A Large-scale Dataset for Robot Design Understanding}

\author{
Tri Le$^{1}$, Toan Nguyen$^{1}$, Quang Tran$^{2}$, Quang Nguyen$^{1}$, Baoru Huang$^{2}$, Hoan Nguyen$^{3}$,\\ Minh Nhat Vu$^{4}$, Tung D. Ta$^{5,6}$, Anh Nguyen$^2$ \\{\small \href{https://airvlab.github.io/robotdesign1m/}{https://airvlab.github.io/robotdesign1m/}} \vspace{-4ex}
\thanks{$^1$ FPT Software AI Center, Vietnam}
\thanks{$^2$ University of Liverpool, UK}
\thanks{$^3$ University     of Information Technology, HCMC, Vietnam} %
\thanks{$^4$ Automation \& Control Institute, TU Wien, Austria}
\thanks{$^5$ Department of Creative Informatics, The University of Tokyo, Japan}%
\thanks{$^{6}$ Faculty of Environment and Information Studies, Keio University, Japan}%
}
\begin{document}

\newtheorem{problem}{Problem}
\newtheorem{lemma}{Lemma}
\newtheorem{theorem}[lemma]{Theorem}
\newtheorem{claim}{Claim}
\newtheorem{corollary}[lemma]{Corollary}
\newtheorem{definition}[lemma]{Definition}
\newtheorem{proposition}[lemma]{Proposition}
\newtheorem{remark}[lemma]{Remark}
\newenvironment{LabeledProof}[1]{\noindent{\it Proof of #1: }}{\qed}

\def\beq#1\eeq{\begin{equation}#1\end{equation}}
\def\bea#1\eea{\begin{align}#1\end{align}}
\def\beg#1\eeg{\begin{gather}#1\end{gather}}
\def\beqs#1\eeqs{\begin{equation*}#1\end{equation*}}
\def\beas#1\eeas{\begin{align*}#1\end{align*}}
\def\begs#1\eegs{\begin{gather*}#1\end{gather*}}

\newcommand{\poly}{\mathrm{poly}}
\newcommand{\eps}{\epsilon}
\newcommand{\e}{\epsilon}
\newcommand{\polylog}{\mathrm{polylog}}
\newcommand{\rob}[1]{\left( #1 \right)} 
\newcommand{\sqb}[1]{\left[ #1 \right]} 
\newcommand{\cub}[1]{\left\{ #1 \right\} } 
\newcommand{\rb}[1]{\left( #1 \right)} 
\newcommand{\abs}[1]{\left| #1 \right|} 
\newcommand{\zo}{\{0, 1\}}
\newcommand{\zonzo}{\zo^n \to \zo}
\newcommand{\zokzo}{\zo^k \to \zo}
\newcommand{\zot}{\{0,1,2\}}
\newcommand{\en}[1]{\marginpar{\textbf{#1}}}
\newcommand{\efn}[1]{\footnote{\textbf{#1}}}
\newcommand{\vecbm}[1]{\boldmath{#1}} 
\newcommand{\uvec}[1]{\hat{\vec{#1}}}
\newcommand{\thv}{\vecbm{\theta}}
\newcommand{\junk}[1]{}
\newcommand{\var}{\mathop{\mathrm{var}}}
\newcommand{\rank}{\mathop{\mathrm{rank}}}
\newcommand{\diag}{\mathop{\mathrm{diag}}}
\newcommand{\tr}{\mathop{\mathrm{tr}}}
\newcommand{\acos}{\mathop{\mathrm{acos}}}
\newcommand{\atantwo}{\mathop{\mathrm{atan2}}}
\newcommand{\SVD}{\mathop{\mathrm{SVD}}}
\newcommand{\quadf}{\mathop{\mathrm{q}}}
\newcommand{\linterp}{\mathop{\mathrm{l}}}
\newcommand{\sgn}{\mathop{\mathrm{sign}}}
\newcommand{\sym}{\mathop{\mathrm{sym}}}
\newcommand{\avg}{\mathop{\mathrm{avg}}}
\newcommand{\mean}{\mathop{\mathrm{mean}}}
\newcommand{\erf}{\mathop{\mathrm{erf}}}
\newcommand{\grad}{\nabla}
\newcommand{\R}{\mathbb{R}}
\newcommand{\defeq}{\triangleq}
\newcommand{\dims}[2]{[#1\!\times\!#2]}
\newcommand{\sdims}[2]{\mathsmaller{#1\!\times\!#2}}
\newcommand{\udims}[3]{#1}
\newcommand{\udimst}[4]{#1}
\newcommand{\com}[1]{\rhd\text{\emph{#1}}}
\newcommand{\ind}{\hspace{1em}}
\newcommand{\argmin}[1]{\underset{#1}{\operatorname{argmin}}}
\newcommand{\floor}[1]{\left\lfloor{#1}\right\rfloor}
\newcommand{\step}[1]{\vspace{0.5em}\noindent{#1}}
\newcommand{\quat}[1]{\ensuremath{\mathring{\mathbf{#1}}}}
\newcommand{\norm}[1]{\left\lVert#1\right\rVert}
\newcommand{\ignore}[1]{}
\newcommand{\specialcell}[2][c]{\begin{tabular}[#1]{@{}c@{}}#2\end{tabular}}
\newcommand*\Let[2]{\State #1 $\gets$ #2}
\newcommand{\algorithmicbreak}{\textbf{break}}
\newcommand{\Break}{\State \algorithmicbreak}
\newcommand{\ra}[1]{\renewcommand{\arraystretch}{#1}}

\renewcommand{\vec}[1]{\mathbf{#1}} 

\algdef{S}[FOR]{ForEach}[1]{\algorithmicforeach\ #1\ \algorithmicdo}
\algnewcommand\algorithmicforeach{\textbf{for each}}
\algrenewcommand\algorithmicrequire{\textbf{Require:}}
\algrenewcommand\algorithmicensure{\textbf{Ensure:}}
\algnewcommand\algorithmicinput{\textbf{Input:}}
\algnewcommand\INPUT{\item[\algorithmicinput]}
\algnewcommand\algorithmicoutput{\textbf{Output:}}
\algnewcommand\OUTPUT{\item[\algorithmicoutput]}

\maketitle
\thispagestyle{empty}
\pagestyle{empty}

\vspace{-1.5ex}

\begin{abstract} 
Robot design is a complex and time-consuming process that requires specialized human expertise. Gaining a deeper understanding of robot design data can enable various applications such as automated design generation, retrieving example designs from the text prompt, or developing AI design assistants. While recent advancements in foundation models present promising approaches to addressing these challenges, progress in this field is hindered by the lack of large-scale robot design datasets. In this paper, we introduce \datasetName{}, a large-scale dataset with 1 million samples. Our dataset features multimodal data collected from scientific literature, covering various robotics domains. We propose a semi-automated data collection pipeline, enabling efficient and diverse data acquisition. To assess the effectiveness of our new \datasetName{} dataset, we conduct extensive experiments across multiple tasks, including design image generation, visual question answering about designs, and design image retrieval. The experimental results demonstrate that \datasetName{} can be used as a challenging benchmark for several robot design understanding tasks. Furthermore, the cross-dataset evaluations demonstrate the ability of \datasetName{} in improving model generalization, proving its data quality. 
\end{abstract}


\section{Introduction} \label{Sec: intro}
Foundation models such as ChatGPT trained on large-scale datasets have demonstrated significant potential across various robotics applications~\cite{gupta2021embodied,ringel2024text2robot}. Recently, several studies have explored the use of large foundation models to support the robot design process, aiming to lower costs, reduce human effort, and push beyond the limits of human creativity~\cite{carbone2022robot, ringel2024text2robot, wang2023diffusebot, li2023evaluation, li2024reinforcement, matthews2023efficient, chan2024creation, hu2022modular}. Despite these advancements, the ability of foundation models to deeply understand and contribute to robot design remains largely underexplored. Designing robots is a highly complex task that integrates knowledge from multiple disciplines and involves a series of critical steps, from functional requirements to physical implementation~\cite{carbone2022robot, spielberg2017functional}. Developing foundation models specifically for robot design has the potential to transform the field by leveraging intelligent, data-driven insights to enhance creativity, efficiency, and performance. In practice, those models can offer insightful feedback, suggest improvements, streamline workflows, and drive innovation~\cite{stella2023can}. 

While foundation models are useful to support the robot design process, the lack of large-scale robot design datasets is a major problem~\cite{hu2023glso, song2024multi}. Efforts have been made to expand datasets for other engineering domains to fuel modern deep learning models~\cite{heyrani2022links,ghezelbash2024mechanical, xu2024cad}. Most of these datasets are synthesized by trying and testing generated parameters through simulation~\cite{picard2023dated, heyrani2022links}. Other datasets~\cite{khan2024text2cad, xu2024cad} provide computer-aided design (CAD) to facilitate research in engineering design generation. Recently, Ghezelbash \etal{}~\cite{ghezelbash2024mechanical} proposes a small-scale dataset comprising mechanical designs and text descriptions. Unlike other generic engineering domains, robot design involves not only the geometric properties of individual components but also their attributes and interrelationships~\cite{makatura2023can}. The absence of specialized large-scale robotic design datasets limits the development of dedicated foundation models tailored for robot design.

\begin{figure}[t]
\centering
\includegraphics[width=0.99\linewidth]{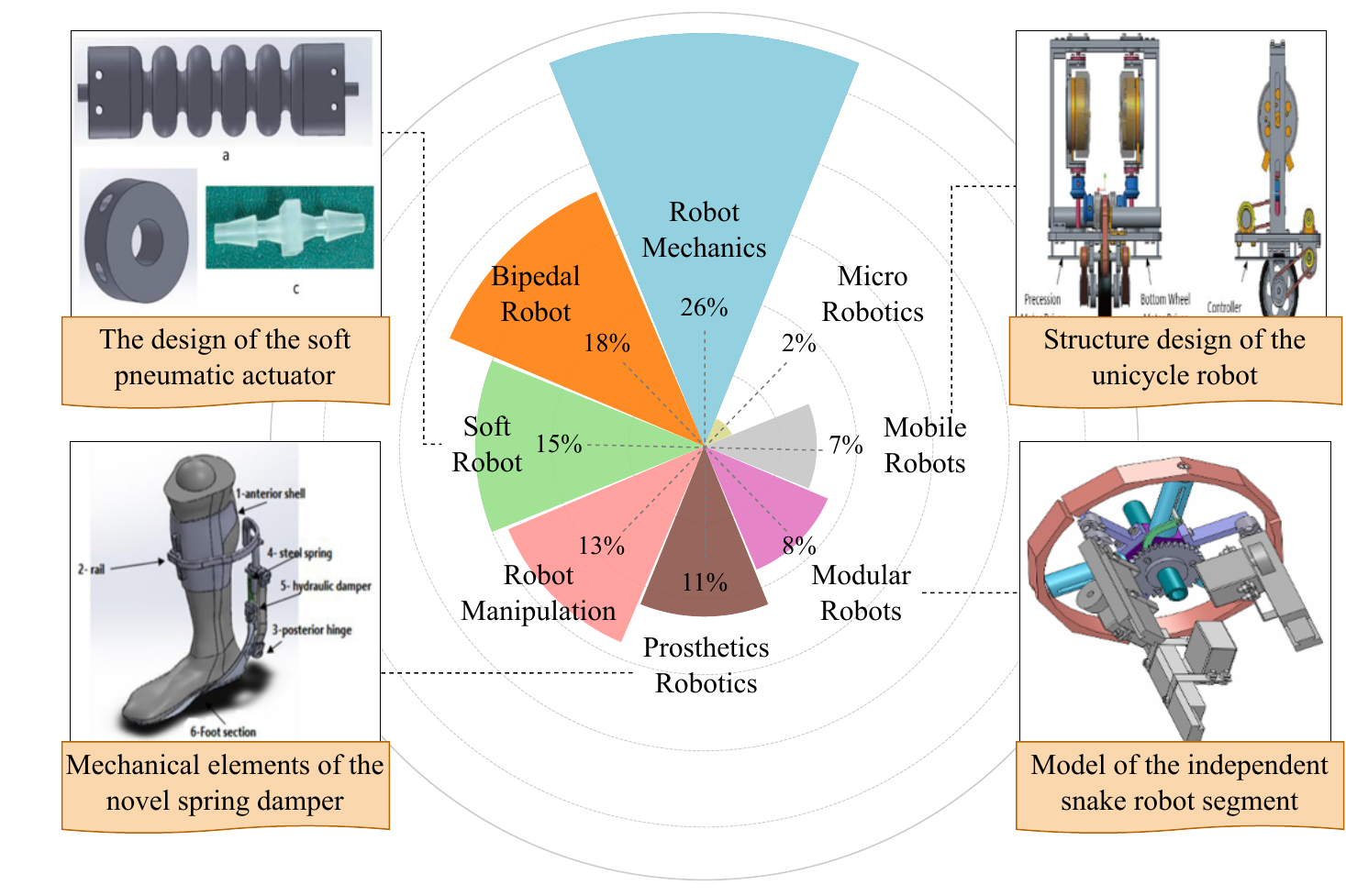}
\vspace{-2ex}
\caption{We introduce \datasetName{}, a new large-scale dataset with 1M samples covering various designs from different robotic disciplines.
}
\label{fig:intro}
\end{figure}

\begin{table*}[!ht]
\centering
\caption{\label{table:design_dataset}Comparison of Design Datasets.}
\vskip 0.1 in
\resizebox{\linewidth}{!}{

\begin{tabular}{lccccccc}
\toprule
\multicolumn{1}{c}{\textbf{Dataset}} & \textbf{$\#$Samples} & \textbf{Design Representation} & \textbf{Domain} & \textbf{Caption} & \textbf{Conversation} & \textbf{Real Design} & \textbf{Public} \\ \midrule
Daele \etal{}~\cite{van2021automated} & 5K & Technical Drawing & Generic & {\color[HTML]{C0C0C0} \ding{56}} & {\color[HTML]{C0C0C0} \ding{56}} & \ding{52} & {\color[HTML]{C0C0C0} \ding{56}} \\
Khan \etal{}~\cite{khan2024fine} & 400 & Technical Drawing & Engineering & {\color[HTML]{C0C0C0} \ding{56}} & {\color[HTML]{C0C0C0} \ding{56}} & \ding{52} & {\color[HTML]{C0C0C0} \ding{56}} \\ 
ABC~\cite{koch2019abc} & 1M & CAD Model & Generic & {\color[HTML]{C0C0C0} \ding{56}} & {\color[HTML]{C0C0C0} \ding{56}} & \ding{52} & \ding{52} \\
DeepCAD~\cite{wu2021deepcad} & 178K & CAD Model & Generic & {\color[HTML]{C0C0C0} \ding{56}} & {\color[HTML]{C0C0C0} \ding{56}} & \ding{52} & \ding{52} \\
Omni-CAD~\cite{xu2024cad} & 453K & CAD Model & Generic & \ding{52} & {\color[HTML]{C0C0C0} \ding{56}} & \ding{52} & \ding{52} \\
CAD Fusion~\cite{wang2025text} & 20K & CAD Model & Generic & \ding{52} & {\color[HTML]{C0C0C0} \ding{56}} & \ding{52} & \ding{52} \\
Text2CAD~\cite{khan2024text2cad} & 170K & CAD Model & Generic & \ding{52} & {\color[HTML]{C0C0C0} \ding{56}} & \ding{52} & \ding{52} \\
Fusion 360 Gallery~\cite{willis2021fusion} & 8K & CAD Model & Generic & {\color[HTML]{C0C0C0} \ding{56}} & {\color[HTML]{C0C0C0} \ding{56}} & \ding{52} & \ding{52} \\
SketchGraph~\cite{seff2020sketchgraphs} & 15M & CAD Sketch & Generic & {\color[HTML]{C0C0C0} \ding{56}} & {\color[HTML]{C0C0C0} \ding{56}} & \ding{52} & \ding{52} \\
Khan \etal{}~\cite{khan2024leveraging} & 300 & CAD Image & Generic & {\color[HTML]{C0C0C0} \ding{56}} & {\color[HTML]{C0C0C0} \ding{56}} & \ding{52} & {\color[HTML]{C0C0C0} \ding{56}} \\
LINKS~\cite{heyrani2022links} & 100M & Parameter & Engineering & {\color[HTML]{C0C0C0} \ding{56}} & {\color[HTML]{C0C0C0} \ding{56}} & {\color[HTML]{C0C0C0} \ding{56}} & \ding{52} \\
Ghezelbash \etal{}~\cite{ghezelbash2024mechanical} & 9K & CAD Image & Engineering & \ding{52} & {\color[HTML]{C0C0C0} \ding{56}} & \ding{52} & \ding{52} \\
Jayanti \etal{}~\cite{jayanti2006developing} & 867 & CAD Model & Engineering & {\color[HTML]{C0C0C0} \ding{56}} & {\color[HTML]{C0C0C0} \ding{56}} & \ding{52} & \ding{52} \\
Lee \etal{}~\cite{lee2022dataset} & 715K & CAD Model & Engineering & {\color[HTML]{C0C0C0} \ding{56}} & {\color[HTML]{C0C0C0} \ding{56}} & {\color[HTML]{C0C0C0} \ding{56}} & \ding{52} \\
\midrule
\textbf{\datasetName{} (ours)} & $\sim1\text{M}$ & \begin{tabular}[c]{@{}c@{}}CAD Image\\ Technical Drawing \end{tabular} & Robotics & \ding{52} & \ding{52} & \ding{52} & \ding{52} \\ 
\bottomrule
\end{tabular}
}
\end{table*}

In practice, collecting a large-scale robot design dataset is a non-trivial task and a time-consuming process. First, collecting technical design data is challenging as robot design involves highly specialized knowledge, and it cannot be generated by existing generative models (e.g., text to images) alone as in~\cite{vuong2023grasp}. While datasets for robot design do exist, they are often proprietary and restricted within the industry, making them inaccessible at scale for the community~\cite{picard2023dated}. Second, collecting large-scale robot design data requires an automated data collection pipeline and a strategy to support the labeling process. In this work, we address these two fundamental challenges. We first propose collecting robot designs from published scientific literature. These sources encompass a wide variety of technical designs of different robotic domains, reflecting extensive research efforts of the community over time. Then, we propose a data creation pipeline with a semi-automated data filtering and labeling process. In addition, we utilize a Large Language Model (LLM) to automatically construct visual instruction-following data, enhancing the utility of the curated data for recent advancements in multi-modal learning~\cite{li2024llara, li2023blip, vuong2024language, bai2023qwen, li2024llava}.

We introduce \textbf{\datasetName{}}, a large-scale dataset for robot design with approximately 1 million (1M) samples. In contrast to prior datasets~\cite{van2021automated, yavartanoo2024text2cad, koch2019abc, wu2021deepcad, ghezelbash2024mechanical, khan2024text2cad, wang2025text}, our dataset consists of 2D drawings and images of robot design and offers two key advantages: \textit{(i)} it is a dedicated dataset for robot design, covering various types of robots and design aspects; \textit{(ii)} it provides high-reliability figures and text sourced from scientific documents. Fig.~\ref{fig:intro} illustrates the robot design topics distribution of the collected scientific documents, along with representative data samples from selected topics. We empirically demonstrate that \datasetName{} enhances the robot design understanding of general-domain models across multiple tasks, including visual question answering, text-image retrieval, and text-to-design image generation. Our findings confirm that general-domain models finetuned on \datasetName{} outperform those finetuned on other related datasets. In summary, our contributions are as follows:
\begin{itemize}
    \item We introduce \datasetName{}, a large-scale dataset dedicated to robot design with over 1M samples and multimodal ground truth.
    \item We intensively benchmark several models on various design-related tasks with \datasetName{}, proving that the quality and diversity of our dataset improve model generalization and validate it as a challenging benchmark for robot design understanding.
\end{itemize}
\section{Related Work} \label{Sec: related_work}
\begin{figure*}[!ht]
    \centering
    \includegraphics[width=\linewidth]{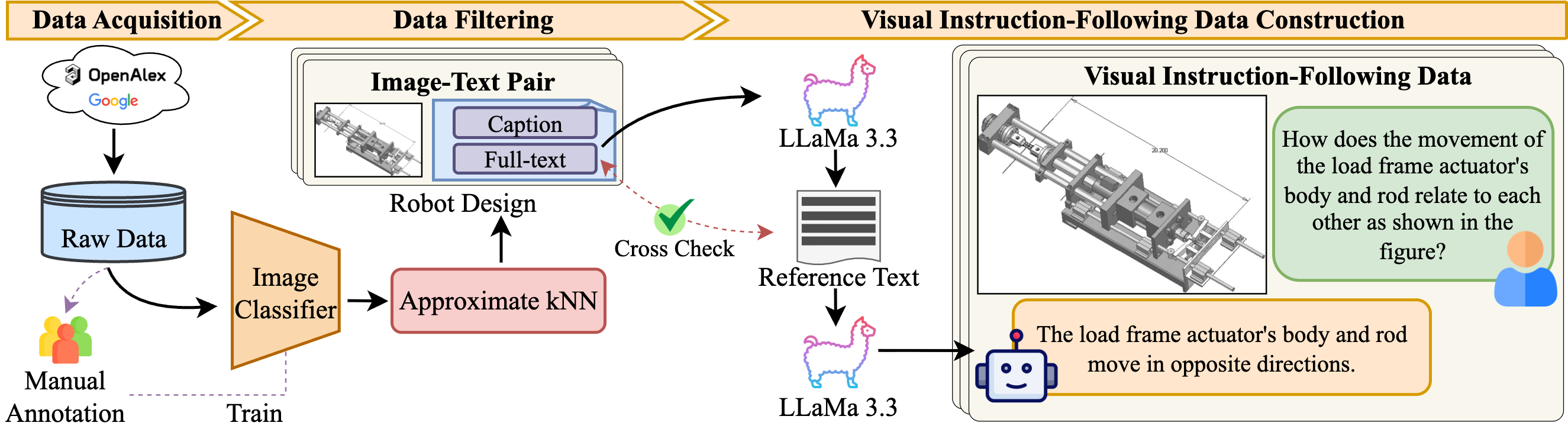}
    \vspace{-1ex}    \caption{\textbf{Dataset creation pipeline.} Our semi-automated dataset creation pipeline comprises three stages: raw data acquisition, data filtering, and visual instruction-following data construction.}
    \label{fig:data-generation}
\end{figure*}

\textbf{Datasets for Robot Design.}
Robotics is inherently interdisciplinary, integrating insights from several fields~\cite{redfield2019definition}. Consequently, a comprehensive robot design dataset must draw from multiple domains~\cite{regenwetter2022deep}. We review existing robot-related design datasets in Table~\ref{table:design_dataset}. Most current design datasets~\cite{van2021automated, schlagenhauf2023text, koch2019abc, wu2021deepcad, xu2024cad, yavartanoo2024text2cad, khan2024text2cad, willis2021fusion, seff2020sketchgraphs, khan2024leveraging, wang2025text} predominantly target generic objects. For example, ABC~\cite{koch2019abc} is a large collection of CAD models in B-rep format and has served as a foundation for later datasets such as DeepCAD~\cite{wu2021deepcad}, Omni-CAD~\cite{xu2024cad}, Text2CAD~\cite{khan2024text2cad}. Although extensive and diverse, these datasets only contain CAD models of generic objects. In contrast, recent works have begun to address the specific needs of mechanical engineering. Lee \etal{}~\cite{lee2022dataset} introduce a large dataset of 3D CAD models featuring mechanical parts. Ghezelbash \etal{}~\cite{ghezelbash2024mechanical} propose a dataset dedicated to mechanisms and gears. However, these datasets remain confined to mechanical engineering, leaving a gap in interdisciplinary robotic domains. Our \datasetName{} fills this gap by collecting large-scale data from multiple robotic fields, for example, soft robots, bipedal robots, etc.

\textbf{Robot Design Automation.}
Numerous approaches have been studied for automating robot design process~\cite{carbone2022robot}, such as evolutionary algorithms~\cite{gupta2021embodied, li2023evaluation}, reinforcement learning~\cite{luck2020data, li2024reinforcement}, and topology optimization~\cite{matthews2023efficient}. Recently, generative models have been leveraged for robot design automation~\cite{chan2024creation, wang2023diffusebot, hu2022modular}. Wang~\etal{}~\cite{wang2023diffusebot} introduce a framework that integrates diffusion models with physics simulation for morphology co-design.
Song~\etal{}~\cite{song2025laser} utilize an LLM within an evolutionary design loop to generate soft robot designs. 
Stella~\etal{}~\cite{stella2023can} leverage LLM to discuss ideas and outline the specifications for the robot design. Jadhav~\etal{}~\cite{jadhav2024large} combine LLM with the finite element method to assess and refine design outcomes. These early studies of LLM applications in robot design inspire our investigation into a potential research direction: creating a large-scale dataset that benefits the fine-tuning process for robot design. Distinct from existing robot design datasets~\cite{gupta2021embodied, li2023evaluation, hu2022modular}, which comprise parameter configurations for specific robot designs, our \datasetName{} dataset includes textual descriptions, 2D drawings, and visual instruction data to advance robot design understanding.

\textbf{Multi-modal Learning.} Recently, multi-modal learning has been widely applied in several tasks~\cite{zong2023self, han2025multimodal}, owing to the significant progress in learning~\cite{li2023blip, ilharco2021openclip, bai2023qwen, li2024llava} and the availability of multi-modal datasets~\cite{pattnayak2024survey, nguyen2025language}. In robot design, several multi-modal learning models are useful when applied in the design process. Text-image retrieval supports design knowledge acquisition, comparative analysis of designs, and inspiration discovery~\cite{kwon2022enabling, song2024multi}. Visual question answering assists engineers in analyzing design choices, providing useful feedback~\cite{stella2023can}, and offers students guidance in their curriculum~\cite{puig2023exploring}. In recognizing these useful applications, we integrate LLM in our pipeline to construct the data that supports multi-modal learning frameworks~\cite{li2024llava}.

\section{The \datasetName{} Dataset} \label{Sec: method}

Fig.~\ref{fig:data-generation} presents an overview of our dataset creation pipeline. We first collect scientific works from the public domain, followed by extracting images and text from these documents. To ensure relevance, we apply a series of semi-automated filtering processes to refine the extracted relevant data. Finally, these images and texts are utilized to construct an instruction-following dataset for further finetuning.

\subsection{Data Acquisition}
We collect scientific documents from multiple public domain sources, including Google Search, OpenAlex~\cite{priem2022openalex}, and open access of IEEE Xplore~\cite{ieeexplore}. To systematically retrieve robot design documents, we first compile a list of robotic keywords extracted from IEEE Robotics \& Automation Letters~\cite{ral2024ral}. This list is expanded using ChatGPT~\cite{openai2024chatgpt}, which generates semantically similar terms and conceptually related keywords. After manual filtering, we finalize a set of 1K keywords, which are then used to query our data sources.
Beyond robotics, we further collect scientific publications from related domains, including mechanical engineering and electrical engineering~\cite{birk2011robotics, redfield2019definition, habib2013engineering}. This data acquisition process results in a corpus of more than 1M documents, covering diverse publication types, including conference proceedings, journal articles, and dissertations.

For content extraction, we utilize PDFFigures~\cite{clark2016pdffigures} to extract images, associated captions, and the full text from the collected documents. This yields a total of 5M image-text pairs, where each image is linked to a caption and the full text of its respective document.

\subsection{Data Filtering}
\textbf{Irrelevant Data Filtering.} The initial extraction includes noisy image-text pairs that are not related to robot design, such as charts, tables, and human-robot experiment images. To filter irrelevant images while maintaining a scalable and automated data collection pipeline, we train an image classifier that is an ensemble model comprising pretrained OpenCLIP~\cite{ilharco2021openclip} and pretrained EfficientNetV2~\cite{tan2021efficientnetv2} as backbones. For the OpenCLIP model, we select the pretrained checkpoints with the highest performance on the ImageNet Sketch benchmark~\cite{wang2019learning}, as this benchmark has the most similar visual concept to robot design schematics. To train the classifier, we \textit{manually annotate} 32,000 images, which are a subset of the extracted data. 
The classifier achieves over 95\% accuracy on our held-out test set before being used to classify the full data. Despite a potential modest amount of noise in our dataset, we demonstrate that its high quality still greatly benefits downstream tasks in our experiments. 

\textbf{Deduplication.} To remove duplicate samples, we use an approximate
k-Nearest-Neighbor (kNN) method as in~\cite{yu2023devil}. Specifically, we compute the embeddings of images using a pretrained OpenCLIP~\cite{ilharco2021openclip} and identify similar images using approximate kNN, implemented by Faiss~\cite{johnson2019billion}. 
Many robot designs are different despite the visual representations being similar. To prevent the erroneous removal of distinct designs, we first identify similar images using approximate kNN. We then compare their associated text descriptions to verify whether they represent the same design. Only samples with both highly similar images and nearly identical textual descriptions are removed. After completing the data filtering process, the final dataset contains approximately 1M image-text pairs relevant to robot design.

\begin{figure*}[htb!]
\centering
\subfigure[Caption length distribution.]{\label{fig:captions_by_length}\includegraphics[width=0.32\linewidth]{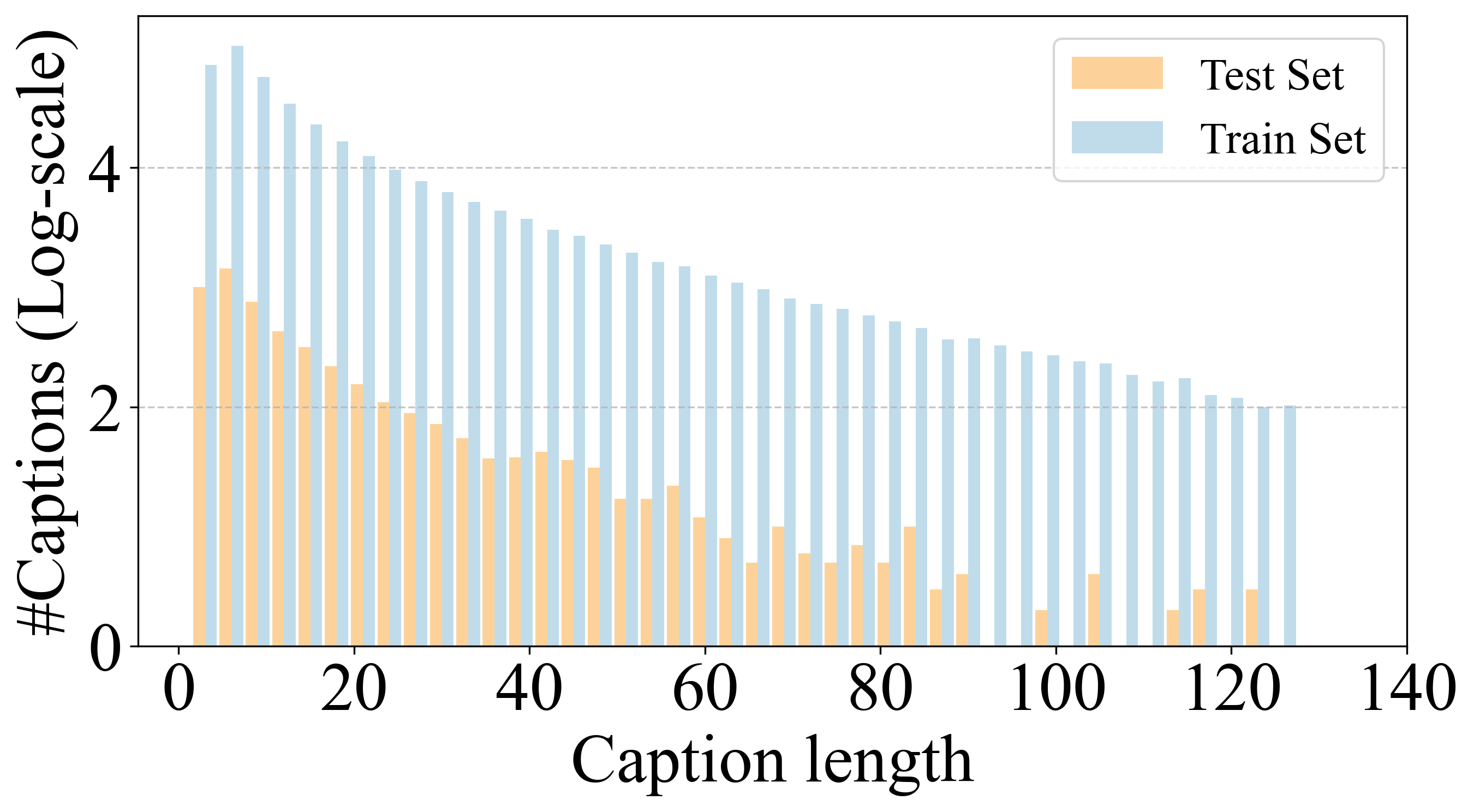}\vspace{-2ex}}
\hspace{0.2ex}
\subfigure[Caption vocabulary.]{\label{fig:caption_vocabulary}\includegraphics[width=0.32\linewidth]{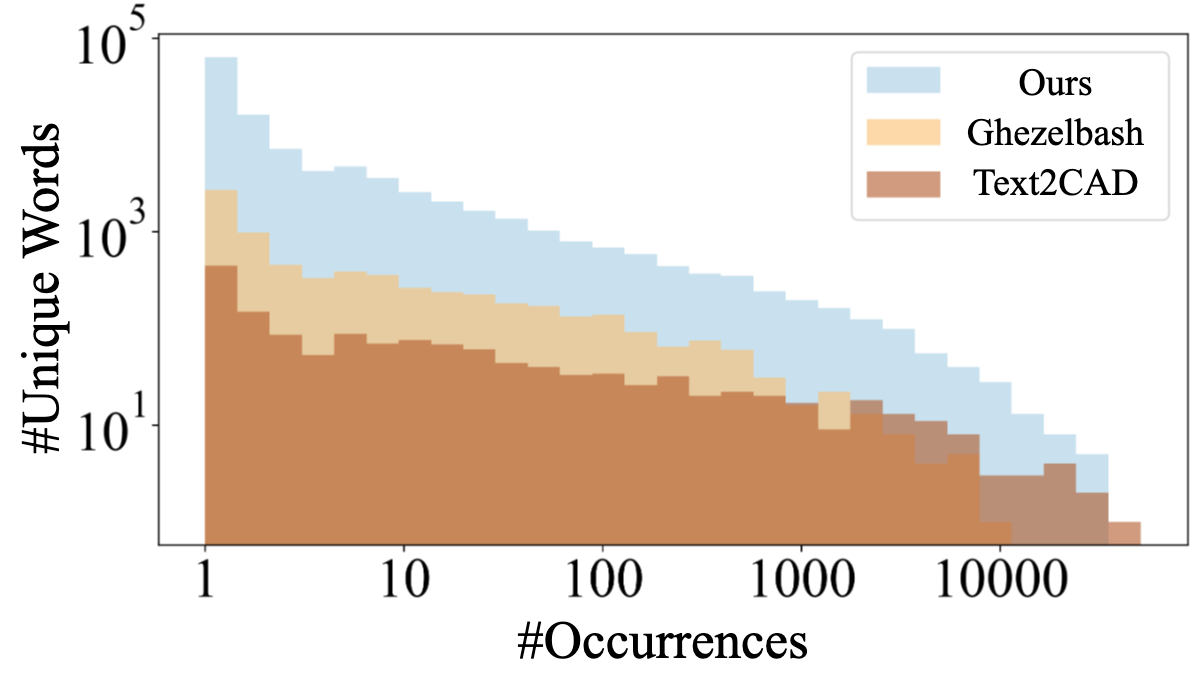}\vspace{-2ex}}
\hspace{0.2ex}
\subfigure[Image keyword distribution.]{\label{fig:keyword_histogram}\includegraphics[width=0.32\linewidth]{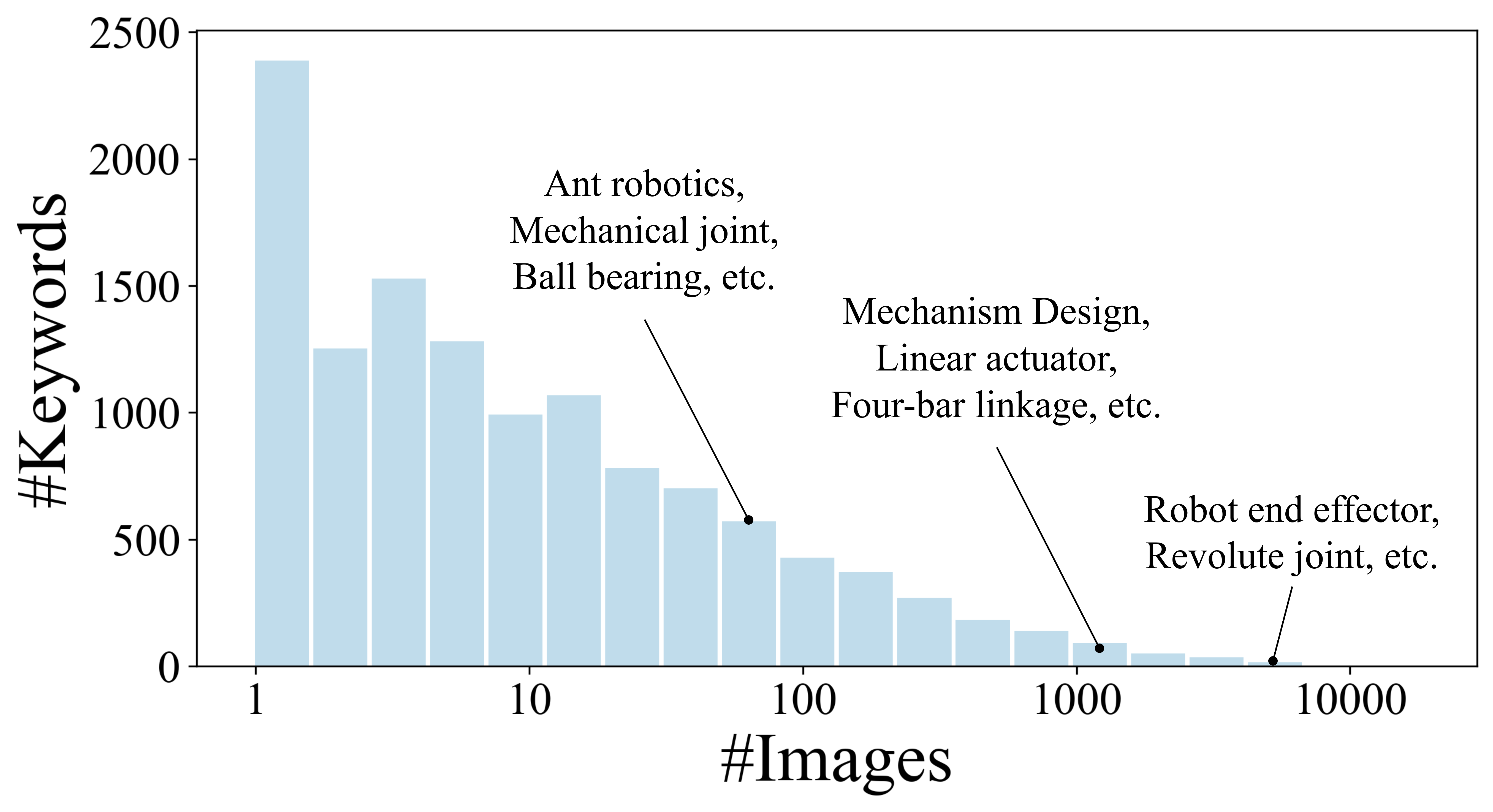}\vspace{-2ex}}
\vspace{1ex}
\caption{\textbf{Dataset Statistics}. We provide statistics on (a) caption length, (b) caption vocabulary, and (c) image keywords.}
\vspace{-1.35ex}
\end{figure*}

\subsection{Visual Instruction-Following Data}

Visual instruction-following data plays a crucial role in finetuning multi-modal learning models~\cite{liu2024visual, bai2023qwen}. However, such data remains scarce in the context of robot design, hindering the development of domain-specific multimodal models. To bridge this gap, we further leverage our curated data to construct a dedicated visual instruction-following dataset for robot design. In this data, each image is associated with a single or multiple question-answer pairs between User and Assistant, which interpret the visual content of the image.

\begin{figure}[htb!]
\centering
\includegraphics[width=1.02\linewidth]{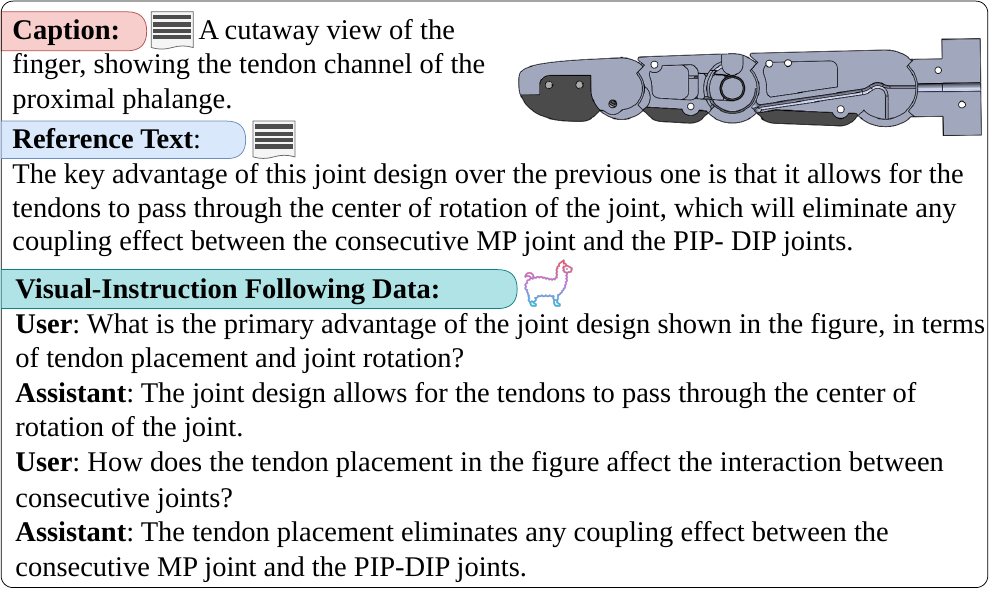}
\vspace{-2ex}
\caption{\textbf{Visual Instruction-Following Data.} Caption and Reference Text are extracted from the documents, while question-answer pairs are generated using LLaMa 3.3 70B~\cite{meta2025llama33}.}
\label{fig:example-data}
\vspace{-2ex}
\end{figure}

\begin{figure*}[h]
    \centering
    \includegraphics[width=0.97\linewidth]{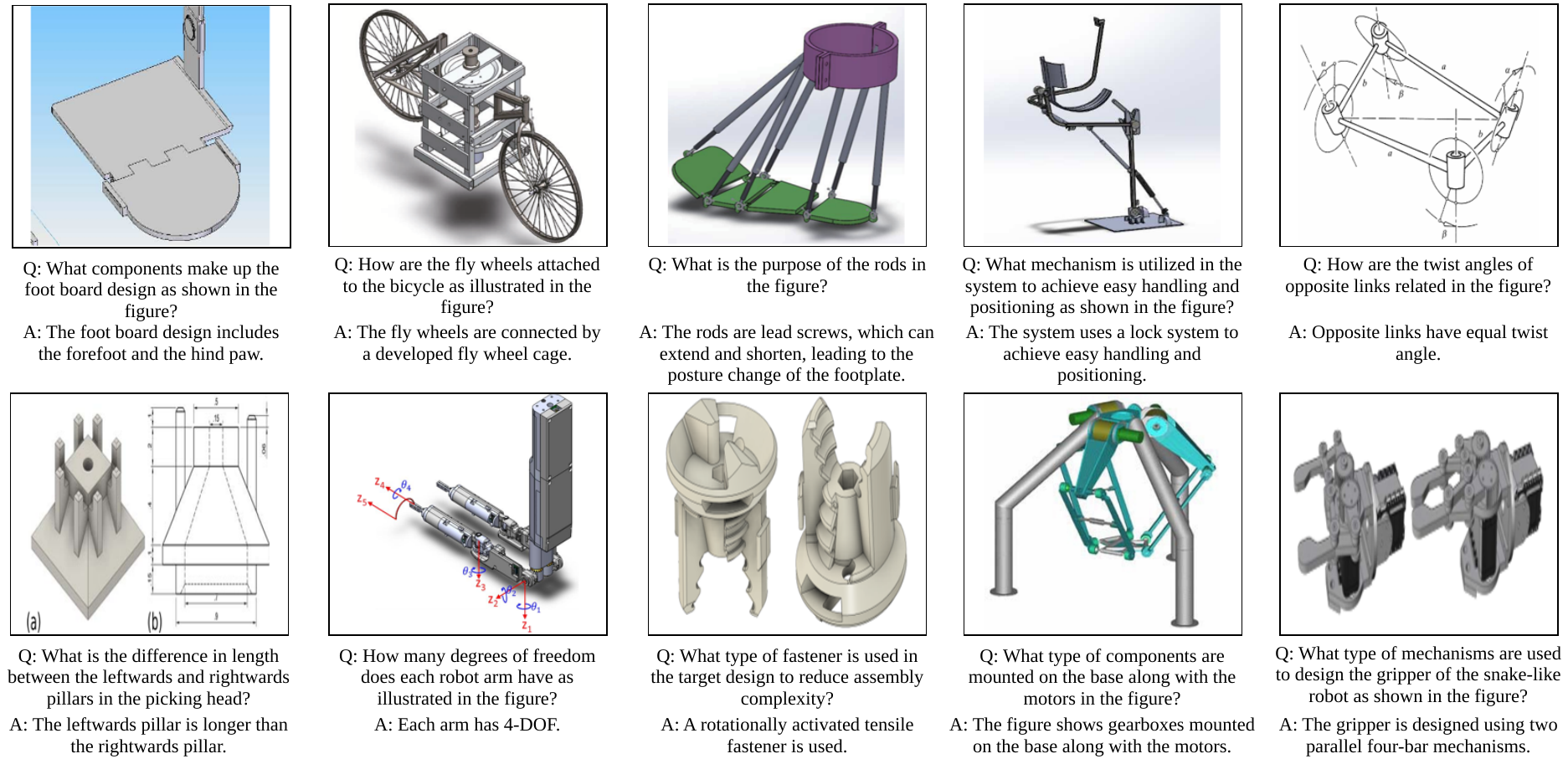}
    \vspace{0.1ex}
    \caption{\textbf{Samples for VQA task from \datasetName{}.}}
    \vspace{-1.5ex}
    \label{fig:samples-vqa}
\end{figure*}

Our visual instruction-following data construction process is designed to enhance reliability while ensuring scalability. First, we observe that several captions provide insufficient descriptions for the images. To enrich the textual content, we leverage an LLM to extract reference texts that are sentences mentioning the image in its respective document. To mitigate LLM hallucination~\cite{zhang2023siren}, the extracted reference texts are validated by cross-checking them with the full text.  Since LLMs are less prone to hallucination when processing short input~\cite{zhang2023siren}, we prompt the LLM to generate question-answer pairs based on the caption and validated reference texts. We invite 5 robotic engineers to manually verify 100 generations before processing the remaining data. In the implementation, we experiment with GPT-4o~\cite{openai2024chatgpt}, LLaMa 3.3 70B~\cite{meta2025llama33}, and Qwen2.5 72B~\cite{yang2024qwen2} and choose LLaMa 3.3 70B as our primary LLM, owing to the balance between affordability and the quality of generation. In addition, to ensure the reference text is extracted accurately, we only provide LLM with paragraphs that contain the image number. Fig.~\ref{fig:example-data} illustrates an example of the caption, reference text, and generated question-answer pairs. We show more samples from our dataset in Fig.~\ref{fig:samples-vqa}. Overall, we construct approximately 1.3 million question-answer pairs associated with the design images.

\subsection{\datasetName{} Statistics}
\textbf{Caption Length Distribution.} \datasetName{} contains captions of varying lengths. Fig.~\ref{fig:captions_by_length} illustrates the distribution of caption lengths in the training and test sets. The average caption length across the dataset is 15.14 words.

\textbf{Caption Vocabulary.} To evaluate caption diversity, we apply lemmatization~\cite{khyani2021interpretation} to map words to their base forms based on context, resulting in a refined caption vocabulary. Fig.~\ref{fig:caption_vocabulary} compares the vocabulary sizes of \datasetName{}, Ghezelbash \etal{}~\cite{ghezelbash2024mechanical}, and Text2CAD~\cite{khan2024text2cad}. \datasetName{} contains 113K unique words, excluding stop words and numbers. Our dataset vocabulary size is approximately 16-78 times higher than Ghezelbash~\cite{ghezelbash2024mechanical} and Text2CAD~\cite{khan2024text2cad}.

\textbf{Visual Content Analysis.} In addition to textual diversity, we assess our dataset’s visual diversity by analyzing the topics and keywords associated with each image. We assign each image the primary topics and keywords from its respective documents, as indexed by OpenAlex~\cite{priem2022openalex}. Fig.~\ref{fig:intro} presents the distribution of robot design topics that contain ``robot'' in their name and their description includes robot design aspects. We shorten the topic names for ease of presentation. The images in \datasetName{} cover 1K topics classified by OpenAlex~\cite{priem2022openalex}. While topics provide a higher-level abstraction, we further examine image keywords to gain a more granular understanding of content diversity. Fig.~\ref{fig:keyword_histogram} visualizes the distribution of image keywords. The images in \datasetName{} are associated with over 12K keywords about robotics and related disciplines, covering a broad range of topics, such as mechanical joints, ball bearings, etc.

\subsection{How will \datasetName{} be used in the community?}
We anticipate several promising directions that can benefit from our dataset. Two direct applications are:
\begin{itemize}
    \item \textit{Robot Design Chatbot}: Finetuning foundation models on visual instruction-following data enables intelligent interaction for robot design. The texts and images in \datasetName{} are sourced from scientific works, ensuring accurate information for chatbot training.
    \item \textit{Cross-modal Design Search}: Cross-modal design search assists engineers and students in looking for design by different modalities such as text and images, facilitating design analogy and inspiration exploration~\cite{kwon2022enabling, song2024multi, li2023deep}. By providing captions and enriched, detailed texts associated with each design, \datasetName{} allows training models that accept various types of queries, e.g., design name, functionality, and appearance.
\end{itemize}

Furthermore, with its large-scale nature, our \datasetName{} dataset can be utilized to develop high-quality data filters~\cite{wang2024finetuned, hessel2021clipscore}, robot design generation~\cite{ringel2024text2robot, wang2023diffusebot}, design retrieval~\cite{kwon2022enabling, song2024multi}, and scientific image captioning~\cite{hsu2021scicap}.

\section{Experiments} \label{Sec: experiments}
To assess the utility of \datasetName{}, we evaluate how models trained on it perform in comparison with those on related datasets on three robot design tasks: Visual question answering on robot design data; Retrieving design images from text; and Generating new designs from user prompts.

\subsection{Visual Question Answering in Robot Design} \label{Sec:vqa}
\textbf{Setup.} In the visual question answering (VQA) task, a model is provided with an image and a textual question and is expected to generate a correct answer. To comprehensively assess the model's understanding of robot design, we construct open-ended question-answer pairs. We use common text generation evaluation metrics~\cite{de2023visual}, specifically BLEU~\cite{papineni2002bleu}, METEOR~\cite{banerjee2005meteor}, and L3Score~\cite{pramanick2024spiqa}. 
For L3Score, we use GPT-4o as the evaluator as in~\cite{pramanick2024spiqa}. 

\textbf{Implementation.} We finetune two pretrained LLM models: LlaVa-1.5~\cite{li2024llava15} and Qwen2-VL~\cite{wang2024qwen2vl} on three datasets: \datasetName{}, Text2CAD~\cite{khan2024text2cad}, and Ghezelbash \etal{}~\cite{ghezelbash2024mechanical} (Ghezelbash). While Text2CAD and Ghezelbash are not specifically designed for robot design, they are the most comparable datasets, as their publicly available data modalities align closely with ours, and their domains focus on engineering design, which shares more similarities with robot design. For Text2CAD, we construct image-text pairs by utilizing multi-view images along with their corresponding generated textual descriptions~\cite{khan2024text2cad}. Since both Text2CAD and Ghezelbash datasets include only image-text pairs, we adapt our pipeline to construct visual instruction-following data for a fair comparison. All models are trained on 4 NVIDIA A100-80GB GPUs until convergence.

\begin{table}[htp]
\centering
\renewcommand\tabcolsep{1pt}
\caption{\label{table:result-vqa} Open-ended Visual Question Answering Results}
\vskip 0.1 in
\resizebox{\linewidth}{!}
{
\begin{tabular}{lc CCC @{\hspace{1em}} CCC @{\hspace{1em}} CCC}
\toprule
& & \multicolumn{3}{c}{\textbf{\datasetName{} (Ours)}} & \multicolumn{3}{c}{\textbf{Text2CAD}~\cite{khan2024text2cad}} & \multicolumn{3}{c}{\textbf{Ghezelbash}~\cite{ghezelbash2024mechanical}} \\ 
\cmidrule(lr){3-5} \cmidrule(lr){6-8} \cmidrule(lr){9-11} 
Models & FT? & B$\uparrow$ & M$\uparrow$ & L$\uparrow$ & B$\uparrow$ & M$\uparrow$ & L$\uparrow$ & B$\uparrow$ & M$\uparrow$ & L$\uparrow$ \\ \midrule
GPT-4o & No & \textbf{2.51} & \textbf{20.83} & 0.30 & 3.12 & 7.74 & 0.30 & \textbf{5.32} & 11.52 & 0.38 \\ \midrule
LlaVa-1.5~\cite{liu2024visual} & No & 0.82 & 4.63 & 0.12 & 3.06 & 6.69 & 0.27 & 1.90 & 7.22 & 0.20 \\
Qwen2-VL~\cite{bai2023qwen} & No & 0.61 & 6.23 & 0.13 & 1.63 & 7.08 & 0.30 & 1.32 & 6.83 & 0.27 \\ \midrule
LlaVa-1.5~\cite{liu2024visual} & Yes & 0.83 & 9.80 & 0.18 & 6.75 & \textbf{32.64} & 0.94 & 3.54 & 17.90 & 0.48 \\
Qwen2-VL~\cite{bai2023qwen} & Yes & 1.32 & 12.33 & \textbf{0.31} & \textbf{6.94} & \textbf{32.64} & \textbf{0.94} & 3.62 & \textbf{18.50} & \textbf{0.52} \\
\bottomrule
\end{tabular}
}
\end{table}

\textbf{VQA Results.} Table~\ref{table:result-vqa} presents the BLEU (B), METEOR (M), and L3Score (L) of the finetuned (FT) models, along with the performance of original pretrained models (without finetuning) and GPT-4o as reference points. The results reveal two key observations. First, general LLMs without finetuning perform poorly across three datasets, highlighting a significant gap in their applicability to engineering and robotic design. While finetuning on domain-specific datasets improves performance, there remains considerable room for further enhancement. 
Second, Table~\ref{table:result-vqa} shows that our \datasetName{} dataset is a \textit{challenging benchmark as models finetuned on our dataset achieve lower accuracy in all metrics, compared to those trained on other datasets}. The low accuracy performance of all models in Table~\ref{table:result-vqa} also indicates that the current methods do not handle well the challenging cases in our dataset. 
However, we note that \textit{models trained on \datasetName{} are neither overfitting to the training set nor suffering from dataset noise}, as evidenced by their superior performance compared to models trained on other datasets in the cross-dataset evaluation (Table~\ref{table:result-compare-dataset}).

\begin{table}[htp]
\vspace{-3ex}
\centering
\renewcommand\tabcolsep{1pt}
\hspace{1ex}
\caption{\label{table:result-compare-dataset} Cross-dataset VQA Generalization Results}
\vskip 0.1 in
\resizebox{\linewidth}{!}
{
\begin{tabular}{r CCC @{\hspace{1em}} CCC @{\hspace{1em}} CCC}
\toprule
\multicolumn{1}{r}{\multirow{2}{*}{\diagbox{Train}{Test}}} & \multicolumn{3}{c}{\textbf{Text2CAD}~\cite{khan2024text2cad}} & \multicolumn{3}{c}{\textbf{Ghezelbash}~\cite{ghezelbash2024mechanical}} & \multicolumn{3}{c}{\textbf{\datasetName{} (Ours)}} \\ 
\cmidrule(lr){2-4} \cmidrule(lr){5-7} \cmidrule(lr){8-10} 
\multicolumn{1}{c}{} & B$\uparrow$ & M$\uparrow$ & L$\uparrow$ & B$\uparrow$ & M$\uparrow$ & L$\uparrow$ & B$\uparrow$ & M$\uparrow$ & L$\uparrow$ \\ 
\midrule
Text2CAD~\cite{khan2024text2cad} & \underline{6.94} & \underline{32.64} & \underline{0.94} & 1.32 & 8.84 & 0.12 & 0.44 & 6.50 & 0.09 \\
Ghezelbash~\cite{ghezelbash2024mechanical} & 2.63 & 15.93 & 0.34 & \underline{3.62} & \underline{18.50} & \underline{0.52} & \textbf{0.84} & \textbf{9.93} & \textbf{0.19} \\ 
\datasetName{} (Ours) & \textbf{3.03} & \textbf{17.42} & \textbf{0.47} & \textbf{2.12} & \textbf{12.93} & \textbf{0.33} & \underline{1.32} & \underline{12.33} & \underline{0.31} \\
\bottomrule
\end{tabular}
}
\end{table}

\begin{figure}[htb!]
\centering
\includegraphics[width=1.02\linewidth]{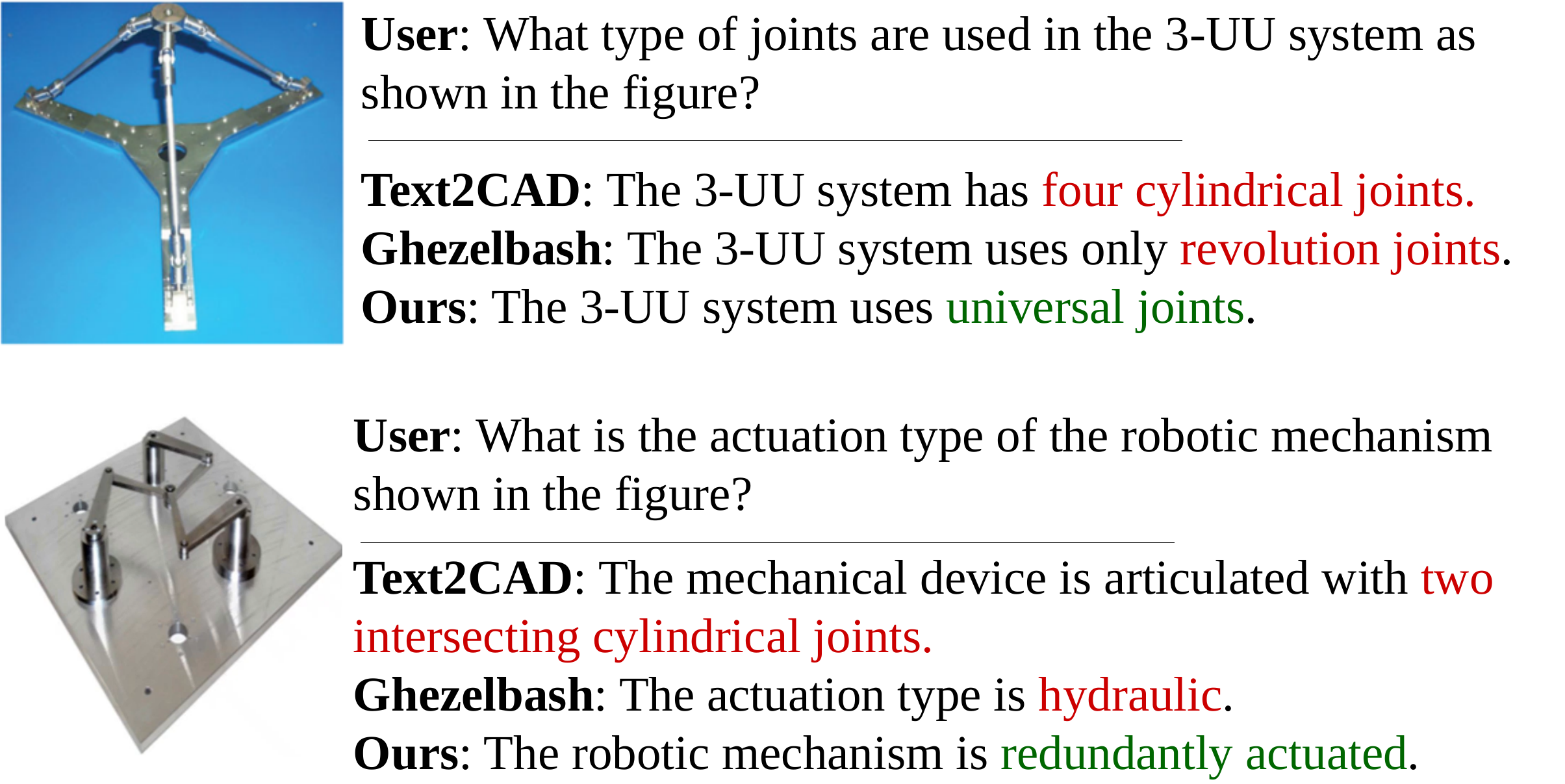}
\vspace{-1ex}
\caption{\textbf{Qualitative VQA Results on Test set.} Answers of Qwen2-VL finetuned on Text2CAD~\cite{khan2024text2cad}, Ghezelbash~\cite{ghezelbash2024mechanical}, and our \datasetName{}. Correct answers are highlighted in \textcolor{ForestGreen}{green}, while incorrect ones are in \textcolor{red}{red}.}
\label{fig:qualitative-vqa}
\vspace{-1.5ex}
\end{figure}

\textbf{Cross-dataset VQA Generalization.} Table~\ref{table:result-compare-dataset} presents the results of finetuning LLMs on a dataset (row) and testing on another dataset (column). For example, the pretrained Qwen2-VL model finetuned on Text2CAD achieves a BLEU score of 0.013 when tested on Ghezelbash. We highlight the best metrics in cross-dataset evaluations in \textbf{bold} and mark in-dataset results with \underline{underlines}. \datasetName{} leads to an improvement of $\approx15$-$62\%$ compared to other datasets in the cross-dataset setting. \textit{This proves the quality of data in \datasetName{} in promoting model generalization.}
Fig.~\ref{fig:qualitative-vqa} provides two qualitative examples from the \datasetName{} test set. In these examples, models finetuned on the Text2CAD and Ghezelbash datasets fail to correctly answer the questions for the given images.

\subsection{Retrieving Design Images from Text}\label{Sec:retrieval}
\vspace{-1ex}
\textbf{Setup.} In text-image retrieval for robot design tasks, a model receives a text query and is tasked with retrieving robot design images that best match the query. We use Recall@$k$ (R@$k$)~\cite{li2024bridging} as the evaluation metric, which measures the proportion of ground-truth instances that appear within the top $k$ retrieved results across all test cases. We report results for $k$ values of 1, 5, and 10.

\textbf{Implementation.} We finetune a state-of-the-art multi-modal encoder BLIP-2~\cite{li2023blip} on three datasets: \datasetName{}, Text2CAD~\cite{khan2024text2cad}, and Ghezelbash~\cite{ghezelbash2024mechanical}. We leverage the image-caption pairs from the selected datasets as training and test data. In this setting, each text query corresponds to a single ground-truth image. All models are trained on 4 NVIDIA A100-80GB GPUs until convergence.

\begin{figure}[h!]
\centering
\includegraphics[width=0.9\linewidth]{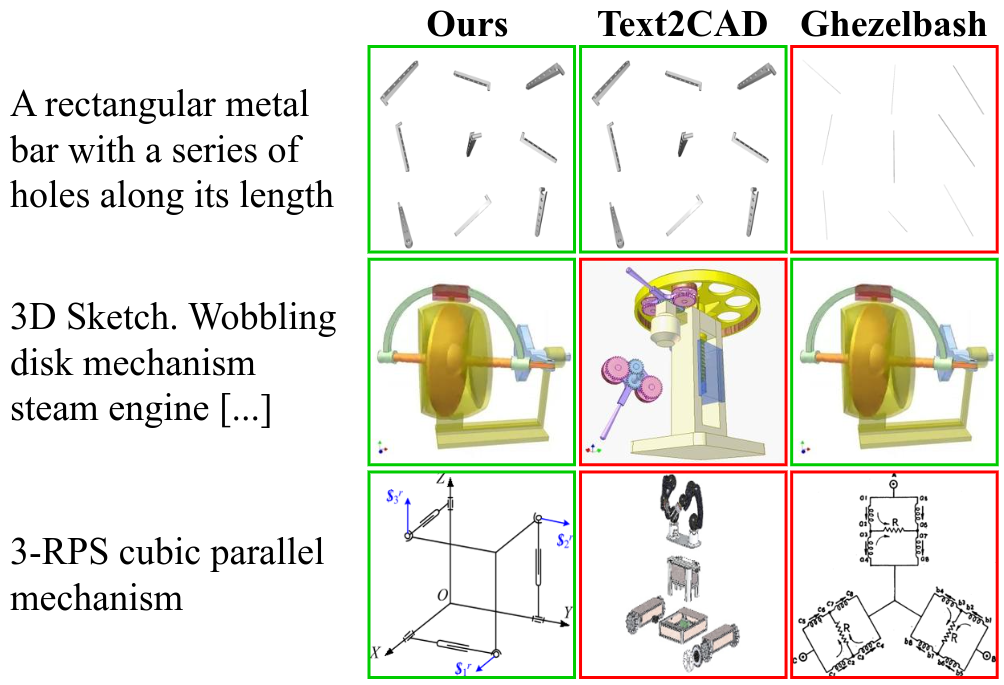}
\vspace{2ex}
\caption{\textbf{Top-1 Text-Image Retrieval Results}. The top-1 retrieval results of models finetuned on three datasets (columns) on three test sets: Text2CAD~\cite{khan2024text2cad} (top), Ghezelbash~\cite{ghezelbash2024mechanical} (middle), and \datasetName{} (bottom). Correct samples are in \textcolor{ForestGreen}{green}, and incorrect ones are in \textcolor{red}{red} boxes.}
\label{fig:qualitative-retrieval}
\vspace{-1ex}
\end{figure}

\begin{table}[h]
\vspace{-4ex}
\centering
\renewcommand\tabcolsep{2.5pt} 
\hspace{1ex}
\caption{\label{table:result-t2i-retrieval} Text-Image Retrieval Results}
\vskip 0.1 in
\resizebox{\linewidth}{!}
{
\begin{tabular}{rccccccccccc}
\toprule
\multirow{2}{*}{\diagbox{Train}{Test}} & \multicolumn{3}{c}{\textbf{Text2CAD}~\cite{khan2024text2cad}} &  & \multicolumn{3}{c}{\textbf{Ghezelbash}~\cite{ghezelbash2024mechanical}} &  & \multicolumn{3}{c}{\textbf{\datasetName{} (Ours)}} \\ \cmidrule{2-4} \cmidrule{6-8} \cmidrule{10-12} 
 & R@1$\uparrow$ & R@5$\uparrow$ & R@10$\uparrow$ &  & R@1$\uparrow$ & R@5$\uparrow$ & R@10$\uparrow$ &  & R@1$\uparrow$ & R@5$\uparrow$ & R@10$\uparrow$ \\ 
\midrule
Text2CAD~\cite{khan2024text2cad} & \underline{16.21} & \underline{40.92} & \underline{53.45} &  & 6.44 & 14.00 & 19.89 &  & \textbf{5.36} & \textbf{10.10} & \textbf{12.98} \\
Ghezelbash~\cite{ghezelbash2024mechanical} & 1.01 & 5.04 & 8.74 &  & \underline{22.11} & \underline{49.44} & \underline{63.00} &  & 4.78 & 9.86 & 13.20 \\
\datasetName{} (Ours) & \textbf{1.68} & \textbf{6.30} & \textbf{9.33} &  & \textbf{8.77} & \textbf{21.11} & \textbf{29.22} &  & \underline{12.9} & \underline{27.46} & \underline{36.28} \\
\bottomrule
\end{tabular}
}
\vspace{-0.5ex}
\end{table}
\textbf{Retrieval Results.} Table~\ref{table:result-t2i-retrieval} presents the finetuning results of the BLIP-2 model on one dataset (rows) and evaluating it on another dataset (columns). Results suggest that \datasetName{} is a challenging benchmark, as models finetuned on it achieve lower performance in all metrics, yet generalize better in cross-dataset testing, compared to those trained on other datasets. Fig.~\ref{fig:qualitative-retrieval} shows top-1 retrieval results for queries from three test sets. The qualitative results show that the model finetuned on \datasetName{} performs well across various design types, including CAD images and drawings.

\subsection{Text-To-Design Image Generation}\label{Sec:image-generation}
\textbf{Setup.} In this task, a model takes a textual description of the robot design as input and generates a corresponding design image. We use the FID~\cite{heusel2017gans} metric, a widely used metric for evaluating text-to-image generation quality~\cite{hartwig2024evaluating}.
FID is used to measure the distance between two distributions of generated images and the source distribution of the ground-truth data. The lower FID is better.

\textbf{Implementation.} We finetune a pretrained state-of-the-art image generation model, Stable Diffusion XL~\cite{rombach2021highresolution} on image-caption pairs of \datasetName{}. We train the model on a server with 4 NVIDIA A100-80GB for 6 days with a batch size of 4 and a gradient accumulation of 8.

\begin{figure}[t]
\centering
\includegraphics[width=0.99\linewidth]{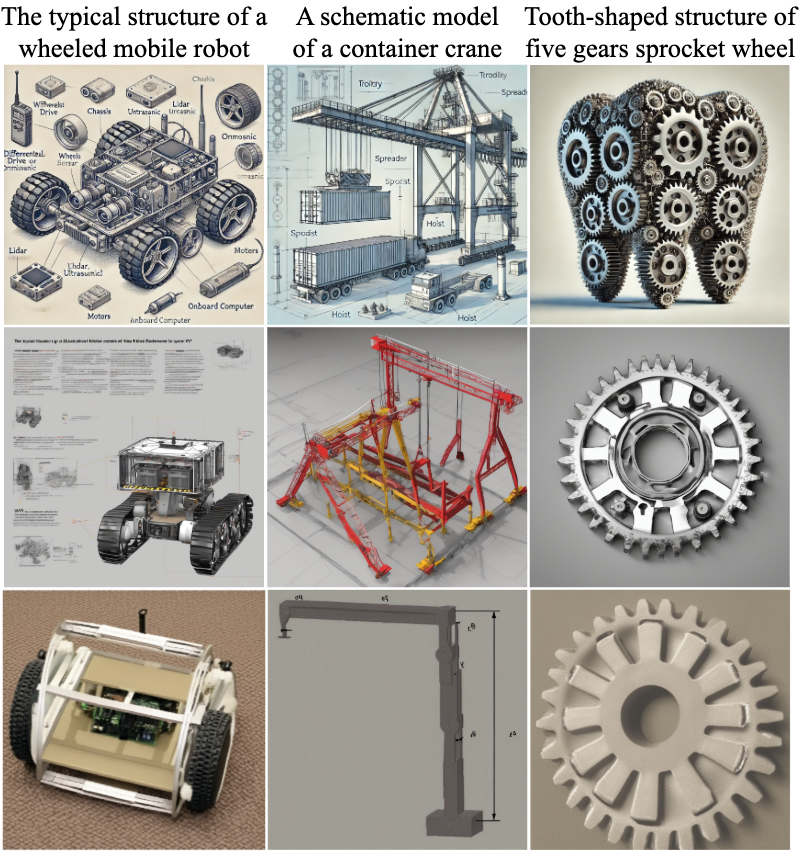}
\vspace{-1ex}
\caption{\textbf{Qualitative Results}. Images generated from the same prompts (column) by GPT-4o (top), the original Stable Diffusion XL (middle), and a finetuned version on \datasetName{} (bottom). The model finetuned on our dataset produces more realistic design images.}
\label{fig:qualitative-t2i}
\end{figure}

\textbf{Generation Results.} We report generation results in Table~\ref{table:result-t2i-generation}. The results show that the model finetuned on \datasetName{} significantly optimizes the FID metric. Fig.~\ref{fig:qualitative-t2i} presents images generated by GPT-4o, a pretrained Stable Diffusion XL model, and the finetuned model on our \datasetName{}. The finetuned model produces more realistic images, while those generated from GPT-4o and the pretrained Stable Diffusion XL focus more on aesthetics.

\begin{table}[htp]
\vspace{-3ex}
\centering
\renewcommand\tabcolsep{20pt} 
\hspace{1ex}
\caption{\label{table:result-t2i-generation} Text-To-Image Generation Results}
\vskip 0.1 in
\resizebox{1\linewidth}{!}
{
\begin{tabular}{ccccc}
\toprule
Model & Finetuned on \datasetName{}? & FID $\downarrow$ \\ 
\midrule
Stable Diffusion XL~\cite{rombach2021highresolution} & No & 45.83 \\
Stable Diffusion XL~\cite{rombach2021highresolution} & Yes & \textbf{39.42} \\ 
\bottomrule
\end{tabular}
}
\end{table}

\begin{figure}[h]
\centering
\includegraphics[width=0.97\linewidth]{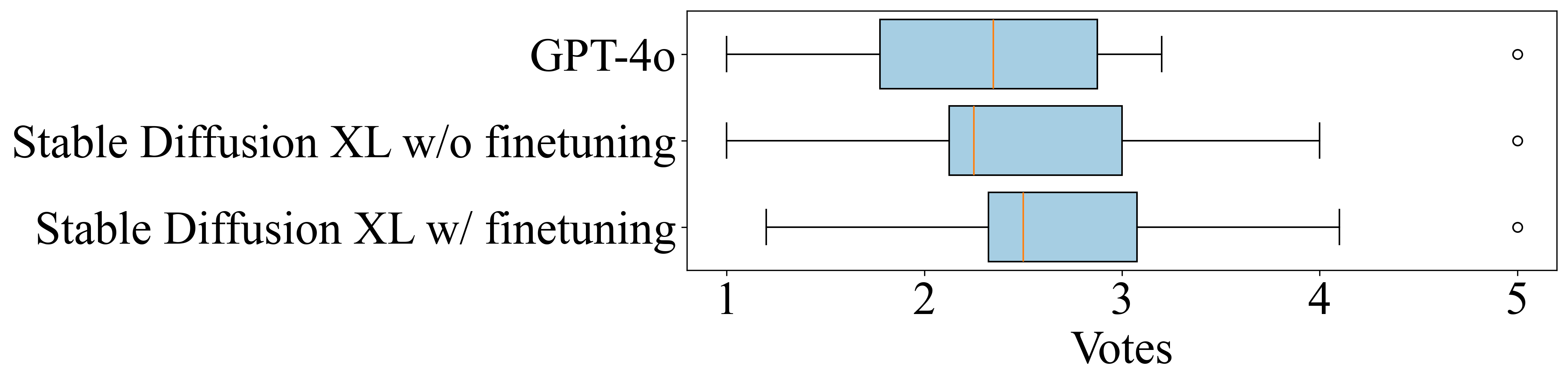}
\vspace{0.5ex}
\caption{\textbf{User Study Results}. Users voted on the design images generated by Stable Diffusion XL~\cite{rombach2021highresolution} with and without finetuning on our dataset.
}
\label{fig:result-user-study}
\end{figure}

\textbf{User Study Results.}
We conducted a user study with 12 robotics engineers and 1,200 evaluations on robot designs generated by GPT-4o, Stable Diffusion XL with and without finetuning on \datasetName{}. 
Participants rated each generated image on a 5-point Likert scale based on its alignment with the input text. Fig.~\ref{fig:result-user-study} presents the results. The model finetuned on \datasetName{} received higher ratings compared to both the original pretrained model and GPT-4o.

\subsection{Discussion}
Through the experiments, we demonstrate the effectiveness of our \datasetName{} in supporting robot design tasks. The diversity of our dataset across various robotic domains enhances the generalization capabilities of multimodal LLMs, as evidenced by its superior performance compared to other datasets in cross-dataset test settings (Table~\ref{table:result-compare-dataset} and Table~\ref{table:result-t2i-retrieval}). While our dataset improves multimodal LLMs in the robotic domain (Table~\ref{table:result-vqa} and Table~\ref{table:result-t2i-generation}), their suboptimal performance highlights the need for further advancements in model architectures to better capture the underlying patterns in robot design data.

\textbf{Limitations.} Although \datasetName{} shows promising results, there are limitations necessitating further studies. 
There is a trade-off between reliability and the richness of textual descriptions. In our dataset, we prioritize reliability and only provide the LLMs with short paragraphs explicitly referring to the images, thereby mitigating hallucinations~\cite{zhang2023siren}. However, this approach inadvertently excludes relevant text that implicitly describes the images. Developing methods to address hallucinations while increasing the richness of extracted texts would enhance the dataset's utility.
In the future, incorporating additional modalities such as parametric designs, CAD models, or converting 2D images into 3D representations~\cite{nguyen2025language, edwards2024sketch2prototype} would provide a more comprehensive understanding of robot design, increasing the usefulness of our dataset. 
\section{Conclusions}\label{Sec: conclusion}
We introduce \textbf{\datasetName{}}, a new large-scale dataset for robot design understanding. Our dataset with 1M samples is constructed by a semi-automated data creation pipeline. \datasetName{} possesses two unique attributes, which are its diversity over a broad range of robotic domains and its high reliability, owing to its data collection sources. The intensive experiments show the usefulness of \datasetName{} as foundation models gain consistent improvements on robot design understanding tasks when finetuned on our dataset than on other related ones.

\bibliographystyle{IEEEtran}
\bibliography{IEEEabrv, reference}

@inproceedings{van2021automated,
  title={An automated engineering assistant: Learning parsers for technical drawings},
  author={Van Daele, Dries and Decleyre, Nicholas and Dubois, Herman and Meert, Wannes},
  booktitle={AAAI},
  year={2021}
}

@inproceedings{vuong2024language,
  title={Language-driven grasp detection},
  author={Vuong, An Dinh and Vu, Minh Nhat and Huang, Baoru and Nguyen, Nghia and Le, Hieu and Vo, Thieu and Nguyen, Anh},
  booktitle={CVPR},
  year={2024}
}

@article{schlagenhauf2023text,
  title={Text detection on technical drawings for the digitization of brown-field processes},
  author={Schlagenhauf, Tobias and Netzer, Markus and Hillinger, Jan},
  journal={Procedia CIRP},
  year={2023},
}

@inproceedings{wu2021deepcad,
  title={Deepcad: A deep generative network for computer-aided design models},
  author={Wu, Rundi and Xiao, Chang and Zheng, Changxi},
  booktitle={ICCV},
  year={2021}
}

@inproceedings{koch2019abc,
  title={Abc: A big cad model dataset for geometric deep learning},
  author={Koch, Sebastian and Matveev, Albert and Jiang, Zhongshi and Williams, Francis and Artemov, Alexey and Burnaev, Evgeny and Alexa, Marc and Zorin, Denis and Panozzo, Daniele},
  booktitle={CVPR},
  year={2019}
}

@article{xu2024cad,
  title={CAD-MLLM: Unifying Multimodality-Conditioned CAD Generation With MLLM},
  author={Xu, Jingwei and Wang, Chenyu and Zhao, Zibo and Liu, Wen and Ma, Yi and Gao, Shenghua},
  journal={arXiv},
  year={2024}
}

@inproceedings{seff2020sketchgraphs,
  title={Sketch{G}raphs: A Large-Scale Dataset for Modeling Relational Geometry in Computer-Aided Design},
  author={Seff, Ari and Ovadia, Yaniv and Zhou, Wenda and Adams, Ryan P.},
  booktitle={ICML Workshop},
  year={2020}
}

@article{khan2024leveraging,
  title={Leveraging Vision-Language Models for Manufacturing Feature Recognition in CAD Designs},
  author={Khan, Muhammad Tayyab and Chen, Lequn and Ng, Ye Han and Feng, Wenhe and Tan, Nicholas Yew Jin and Moon, Seung Ki},
  journal={JCISE},
  year={2025}
}

@article{stella2023can,
  title={How can LLMs transform the robotic design process?},
  author={Stella, Francesco and Della Santina, Cosimo and Hughes, Josie},
  journal={Nature machine intelligence},
  year={2023},
}

@article{makatura2023can,
  title={How Can Large Language Models Help Humans in Design and Manufacturing?},
  author={Makatura, Liane and Foshey, Michael and Wang, Bohan and H{\"a}hnLein, Felix and Ma, Pingchuan and Deng, Bolei and Tjandrasuwita, Megan and Spielberg, Andrew and Owens, Crystal Elaine and Chen, Peter Yichen and others},
  journal={arXiv},
  year={2023}
}

@article{li2024llava,
  title={Llava-med: Training a large language-and-vision assistant for biomedicine in one day},
  author={Li, Chunyuan and Wong, Cliff and Zhang, Sheng and Usuyama, Naoto and Liu, Haotian and Yang, Jianwei and Naumann, Tristan and Poon, Hoifung and Gao, Jianfeng},
  journal={NeurIPS},
  year={2024}
}

@article{priem2022openalex,
  title={OpenAlex: A fully-open index of scholarly works, authors, venues, institutions, and concepts},
  author={Priem, Jason and Piwowar, Heather and Orr, Richard},
  journal={arXiv},
  year={2022}
}

@misc{openai2024chatgpt,
  author = {OpenAI},
  title = {{Hello GPT-4o}},
  howpublished = {Software},
  url = {https://openai.com/index/hello-gpt-4o/},
  note = {Accessed: May 12th 2025.},
}

@inproceedings{clark2016pdffigures,
  title={Pdffigures 2.0: Mining figures from research papers},
  author={Clark, Christopher and Divvala, Santosh},
  booktitle={Proceedings of the 16th ACM/IEEE-CS on Joint Conference on Digital Libraries},
  year={2016}
}

@software{ilharco2021openclip,
  author       = {Ilharco, Gabriel and
                  Wortsman, Mitchell and
                  Wightman, Ross and
                  Gordon, Cade and
                  Carlini, Nicholas and
                  Taori, Rohan and
                  Dave, Achal and
                  Shankar, Vaishaal and
                  Namkoong, Hongseok and
                  Miller, John and
                  Hajishirzi, Hannaneh and
                  Farhadi, Ali and
                  Schmidt, Ludwig},
  title        = {OpenCLIP},
  year         = 2021,
}

@inproceedings{wang2019learning,
        title={Learning Robust Global Representations by Penalizing Local Predictive Power},
        author={Wang, Haohan and Ge, Songwei and Lipton, Zachary and Xing, Eric P},
        booktitle={NeurIPS},
        year={2019}
}

@article{regenwetter2022deep,
  title={Deep generative models in engineering design: A review},
  author={Regenwetter, Lyle and Nobari, Amin Heyrani and Ahmed, Faez},
  journal={Journal of Mechanical Design},
  year={2022},
}

@inproceedings{ringel2024text2robot,
  title={Text2robot: Evolutionary robot design from text descriptions},
  author={Ringel, Ryan P and Charlick, Zachary S and Liu, Jiaxun and Xia, Boxi and Chen, Boyuan},
  booktitle={ICRA},
  year={2025},
}

@article{wang2023diffusebot,
  title={Diffusebot: Breeding soft robots with physics-augmented generative diffusion models},
  author={Wang, Tsun-Hsuan Johnson and Zheng, Juntian and Ma, Pingchuan and Du, Yilun and Kim, Byungchul and Spielberg, Andrew and Tenenbaum, Josh and Gan, Chuang and Rus, Daniela},
  journal={NeurIPS},
  year={2023}
}

@inproceedings{tan2021efficientnetv2,
  title={Efficientnetv2: Smaller models and faster training},
  author={Tan, Mingxing and Le, Quoc},
  booktitle={ICML},
  year={2021},
}

@article{liu2024visual,
  title={Visual instruction tuning},
  author={Liu, Haotian and Li, Chunyuan and Wu, Qingyang and Lee, Yong Jae},
  journal={NeurIPS},
  year={2024}
}

@article{wang2024finetuned,
  title={Finetuned Multimodal Language Models Are High-Quality Image-Text Data Filters},
  author={Wang, Weizhi and Mrini, Khalil and Yang, Linjie and Kumar, Sateesh and Tian, Yu and Yan, Xifeng and Wang, Heng},
  journal={arXiv},
  year={2024}
}

@inproceedings{hsu2021scicap,
  title={SciCap: Generating captions for scientific figures},
  author={Hsu, Ting-Yao and Giles, C Lee and Huang, Ting-Hao},
  booktitle={EMNLP Findings},
  year={2021}
}

@inproceedings{hessel2021clipscore,
  title={Clipscore: A reference-free evaluation metric for image captioning},
  author={Hessel, Jack and Holtzman, Ari and Forbes, Maxwell and Le Bras, Ronan and Choi, Yejin},
  booktitle={EMNLP},
  year={2021}
}

@article{johnson2019billion,
  title={Billion-scale similarity search with {GPUs}},
  author={Johnson, Jeff and Douze, Matthijs and J{\'e}gou, Herv{\'e}},
  journal={IEEE Transactions on Big Data},
  year={2019},
}

@article{bai2023qwen,
  title={Qwen-vl: A versatile vision-language model for understanding, localization, text reading, and beyond},
  author={Bai, Jinze and Bai, Shuai and Yang, Shusheng and Wang, Shijie and Tan, Sinan and Wang, Peng and Lin, Junyang and Zhou, Chang and Zhou, Jingren},
  journal={arXiv},
  year={2023}
}

@inproceedings{vuong2023grasp,
  title={Grasp-anything: Large-scale grasp dataset from foundation models},
  author={Vuong, An Dinh and Vu, Minh Nhat and Le, Hieu and Huang, Baoru and Binh, Huynh Thi Thanh and Vo, Thieu and Kugi, Andreas and Nguyen, Anh},
  booktitle={ICRA},
  year={2024}
}

@inproceedings{nguyen2025language,
  title={Language-driven 6-dof grasp detection using negative prompt guidance},
  author={Nguyen, Toan and Vu, Minh Nhat and Huang, Baoru and Vuong, An and Vuong, Quan and Le, Ngan and Vo, Thieu and Nguyen, Anh},
  booktitle={ECCV},
  year={2024},
}

@article{de2023visual,
  title={Visual question answering: A survey on techniques and common trends in recent literature},
  author={de Faria, Ana Cl{\'a}udia Akemi Matsuki and Bastos, Felype de Castro and da Silva, Jos{\'e} Victor Nogueira Alves and Fabris, Vitor Lopes and Uchoa, Valeska de Sousa and Neto, D{\'e}cio Gon{\c{c}}alves de Aguiar and Santos, Claudio Filipi Goncalves dos},
  journal={arXiv},
  year={2023}
}

@inproceedings{hu2023glso,
  title={GLSO: grammar-guided latent space optimization for sample-efficient robot design automation},
  author={Hu, Jiaheng and Whitman, Julian and Choset, Howie},
  booktitle={CoRL},
  year={2022},
}

@article{ghezelbash2024mechanical,
  title={A dataset for mechanical mechanisms},
  author={Ghezelbash, Farshid and Eskandari, Amir Hossein and Bidhendi, Amir J},
  journal={arXiv},
  year={2024}
}

@inproceedings{heyrani2022links,
  title={Links: A dataset of a hundred million planar linkage mechanisms for data-driven kinematic design},
  author={Heyrani Nobari, Amin and Srivastava, Akash and Gutfreund, Dan and Ahmed, Faez},
  booktitle={IDETC-CIE},
  year={2022},
}

@inproceedings{picard2023dated,
  title={Dated: Guidelines for creating synthetic datasets for engineering design applications},
  author={Picard, Cyril and Schiffmann, J{\"u}rg and Ahmed, Faez},
  booktitle={IDETC-CIE},
  year={2023},
}

@article{jayanti2006developing,
  title={Developing an engineering shape benchmark for CAD models},
  author={Jayanti, Subramaniam and Kalyanaraman, Yagnanarayanan and Iyer, Natraj and Ramani, Karthik},
  journal={Computer-Aided Design},
  year={2006},
}

@article{yavartanoo2024text2cad,
  title={Text2CAD: Text to 3D CAD Generation via Technical Drawings},
  author={Yavartanoo, Mohsen and Hong, Sangmin and Neshatavar, Reyhaneh and Lee, Kyoung Mu},
  journal={arXiv},
  year={2024}
}

@article{khan2024text2cad,
  title={Text2cad: Generating sequential cad designs from beginner-to-expert level text prompts},
  author={Khan, Mohammad S and Sinha, Sankalp and Sheikh, Talha U and Stricker, Didier and Ali, Sk A and Afzal, Muhammad Z},
  journal={NeurIPS},
  year={2024}
}

@article{lee2022dataset,
  title={Dataset and method for deep learning-based reconstruction of 3D CAD models containing machining features for mechanical parts},
  author={Lee, Hyunoh and Lee, Jinwon and Kim, Hyungki and Mun, Duhwan},
  journal={Journal of Computational Design and Engineering},
  year={2022},
}

@article{willis2021fusion,
  title={Fusion 360 gallery: A dataset and environment for programmatic cad construction from human design sequences},
  author={Willis, Karl DD and Pu, Yewen and Luo, Jieliang and Chu, Hang and Du, Tao and Lambourne, Joseph G and Solar-Lezama, Armando and Matusik, Wojciech},
  journal={ACM Transactions on Graphics (TOG)},
  year={2021},
}

@inproceedings{khan2024fine,
  title={Fine-tuning vision-language model for automated engineering drawing information extraction},
  author={Khan, Muhammad Tayyab and Chen, Lequn and Ng, Ye Han and Feng, Wenhe and Tan, Nicholas Yew Jin and Moon, Seung Ki},
  booktitle={ICIAI},
  year={2025}
}

@misc{ral2024ral,
  author = {{IEEE Robotics {\&} Automation Society}},
  title = {{RA-L Keywords}},
  howpublished = {Software},
  url = {https://www.ieee-ras.org/publications/ra-l/keywords},
  note = {Accessed: May 12th 2025.},
}

@article{birk2011robotics,
  title={What is robotics? An interdisciplinary field is getting even more diverse},
  author={Birk, Andreas},
  journal={IEEE robotics \& automation magazine},
  year={2011}
}

@article{redfield2019definition,
  title={A definition for robotics as an academic discipline},
  author={Redfield, Signe},
  journal={Nature Machine Intelligence},
  year={2019},
}

@inproceedings{papineni2002bleu,
  title={Bleu: a method for automatic evaluation of machine translation},
  author={Papineni, Kishore and Roukos, Salim and Ward, Todd and Zhu, Wei-Jing},
  booktitle={ACL},
  year={2002}
}

@book{habib2013engineering,
  title={Engineering creative design in robotics and mechatronics},
  author={Habib, Maki K and Davim, J Paulo},
  year={2013},
}

@inproceedings{puig2023exploring,
  title={Exploring the Application of ChatGPT in Mechanical Engineering Education},
  author={Puig Ortiz, Joan and P{\`a}mies Vil{\`a}, Rosa and Jordi Nebot, Llu{\"\i}sa},
  booktitle={SEFI Annual Conference Annual Conference},
  year={2023}
}

@inproceedings{li2023blip,
  title={Blip-2: Bootstrapping language-image pre-training with frozen image encoders and large language models},
  author={Li, Junnan and Li, Dongxu and Savarese, Silvio and Hoi, Steven},
  booktitle={ICML},
  year={2023},
}

@article{li2024bridging,
  title={Bridging Modalities: A Survey of Cross-Modal Image-Text Retrieval},
  author={Li, Tieying and Kong, Lingdu and Yang, Xiaochun and Wang, Bin and Xu, Jiaxing},
  journal={Chinese Journal of Information Fusion},
  year={2024},
}

@book{carbone2022robot,
  title={Robot Design: From Theory to Service Applications},
  author={Carbone, Giuseppe and Laribi, Med Amine},
  year={2022},
  publisher={Springer Nature}
}

@inproceedings{spielberg2017functional,
  title={Functional co-optimization of articulated robots},
  author={Spielberg, Andrew and Araki, Brandon and Sung, Cynthia and Tedrake, Russ and Rus, Daniela},
  booktitle={ICRA},
  year={2017},
}

@article{pramanick2024spiqa,
  title={Spiqa: A dataset for multimodal question answering on scientific papers},
  author={Pramanick, Shraman and Chellappa, Rama and Venugopalan, Subhashini},
  journal={NeurIPS},
  year={2024}
}

@inproceedings{rombach2021highresolution,
  title={High-resolution image synthesis with latent diffusion models},
  author={Rombach, Robin and Blattmann, Andreas and Lorenz, Dominik and Esser, Patrick and Ommer, Bj{\"o}rn},
  booktitle={CVPR},
  year={2022}
}

@article{heusel2017gans,
  title={Gans trained by a two time-scale update rule converge to a local nash equilibrium},
  author={Heusel, Martin and Ramsauer, Hubert and Unterthiner, Thomas and Nessler, Bernhard and Hochreiter, Sepp},
  journal={NeurIPS},
  year={2017}
}

@article{hartwig2024evaluating,
  title={Evaluating Text to Image Synthesis: Survey and Taxonomy of Image Quality Metrics},
  author={Hartwig, Sebastian and Engel, Dominik and Sick, Leon and Kniesel, Hannah and Payer, Tristan and Ropinski, Timo and others},
  journal={arXiv},
  year={2024}
}

@inproceedings{banerjee2005meteor,
  title={METEOR: An automatic metric for MT evaluation with improved correlation with human judgments},
  author={Banerjee, Satanjeev and Lavie, Alon},
  booktitle={ACL Workshop},
  year={2005}
}

@article{yu2023devil,
  title={The devil is in the details: A deep dive into the rabbit hole of data filtering},
  author={Yu, Haichao and Tian, Yu and Kumar, Sateesh and Yang, Linjie and Wang, Heng},
  journal={arXiv},
  year={2023}
}

@inproceedings{hu2022modular,
  title={Modular robot design optimization with generative adversarial networks},
  author={Hu, Jiaheng and Whitman, Julian and Travers, Matthew and Choset, Howie},
  booktitle={ICRA},
  year={2022},
}

@article{chan2024creation,
  title={Creation of Novel Soft Robot Designs using Generative AI},
  author={Chan, Wee Kiat and Wang, PengWei and Yeow, Raye Chen-Hua},
  journal={arXiv},
  year={2024}
}

@article{jadhav2024large,
  title={Large language model agent as a mechanical designer},
  author={Jadhav, Yayati and Barati Farimani, Amir},
  journal={Journal of Engineering Design},
  year={2026}
}

@inproceedings{song2025laser,
  title={Laser: Towards diversified and generalizable robot design with large language models},
  author={Song, Junru and Yang, Yang and Xiao, Huan and Peng, Wei and Yao, Wen and Wang, Feifei},
  booktitle={ICLR},
  year={2025}
}

@inproceedings{li2024reinforcement,
  title={Reinforcement learning for freeform robot design},
  author={Li, Muhan and Matthews, David and Kriegman, Sam},
  booktitle={ICRA},
  year={2024},
}

@article{gupta2021embodied,
  title={Embodied intelligence via learning and evolution},
  author={Gupta, Agrim and Savarese, Silvio and Ganguli, Surya and Fei-Fei, Li},
  journal={Nature communications},
  year={2021},
}

@article{li2023evaluation,
  title={Evaluation of frameworks that combine evolution and learning to design robots in complex morphological spaces},
  author={Li, Wei and Buchanan, Edgar and Le Goff, L{\'e}ni K and Hart, Emma and Hale, Matthew F and Wei, Bingsheng and De Carlo, Matteo and Angus, Mike and Woolley, Robert and Gan, Zhongxue and others},
  journal={IEEE Transactions on Evolutionary Computation},
  year={2023},
}

@inproceedings{luck2020data,
  title={Data-efficient co-adaptation of morphology and behaviour with deep reinforcement learning},
  author={Luck, Kevin Sebastian and Amor, Heni Ben and Calandra, Roberto},
  booktitle={CoRL},
  year={2019},
}

@article{matthews2023efficient,
  title={Efficient automatic design of robots},
  author={Matthews, David and Spielberg, Andrew and Rus, Daniela and Kriegman, Sam and Bongard, Josh},
  journal={Proceedings of the National Academy of Sciences},
  year={2023},
}

@article{zong2023self,
  title={Self-supervised multimodal learning: A survey},
  author={Zong, Yongshuo and Mac Aodha, Oisin and Hospedales, Timothy M},
  journal={TPAMI},
  year={2024}
}

@article{song2024multi,
  title={Multi-modal machine learning in engineering design: A review and future directions},
  author={Song, Binyang and Zhou, Rui and Ahmed, Faez},
  journal={JCISE},
  year={2024},
}

@article{kwon2022enabling,
  title={Enabling multi-modal search for inspirational design stimuli using deep learning},
  author={Kwon, Elisa and Huang, Forrest and Goucher-Lambert, Kosa},
  journal={AI EDAM},
  year={2022},
}

@article{li2023deep,
  title={Deep learning methods of cross-modal tasks for conceptual design of product shapes: A review},
  author={Li, Xingang and Wang, Ye and Sha, Zhenghui},
  journal={Journal of Mechanical Design},
  year={2023},
}

@article{khyani2021interpretation,
  title={An interpretation of lemmatization and stemming in natural language processing},
  author={Khyani, Divya and Siddhartha, BS and Niveditha, NM and Divya, BM},
  journal={Journal of University of Shanghai for Science and Technology},
  year={2021}
}

@article{zhang2023siren,
  title={Siren's song in the AI ocean: a survey on hallucination in large language models},
  author={Zhang, Yue and Li, Yafu and Cui, Leyang and Cai, Deng and Liu, Lemao and Fu, Tingchen and Huang, Xinting and Zhao, Enbo and Zhang, Yu and Chen, Yulong and others},
  journal={Computational Linguistics},
  year={2025},
}

@misc{meta2025llama33,
  author = {Meta},
  title = {{Llama 3.3 - Model Cards \& Prompt formats}},
  howpublished = {Software},
  url = {https://www.llama.com/docs/model-cards-and-prompt-formats/llama3_3/},
  note = {Accessed: May 12th 2025.},
}

@article{yang2024qwen2,
  title={Qwen2. 5 technical report},
  author={Yang, An and Yang, Baosong and Zhang, Beichen and Hui, Binyuan and Zheng, Bo and Yu, Bowen and Li, Chengyuan and Liu, Dayiheng and Huang, Fei and Wei, Haoran and others},
  journal={arXiv},
  year={2024}
}

@article{edwards2024sketch2prototype,
  title={Sketch2Prototype: rapid conceptual design exploration and prototyping with generative AI},
  author={Edwards, Kristen M and Man, Brandon and Ahmed, Faez},
  journal={Proceedings of the Design Society},
  year={2024},
}

@inproceedings{li2024llara,
  title={LLaRA: Supercharging Robot Learning Data for Vision-Language Policy},
  author={Li, Xiang and Mata, Cristina and Park, Jongwoo and Kahatapitiya, Kumara and Jang, Yoo Sung and Shang, Jinghuan and Ranasinghe, Kanchana and Burgert, Ryan D and Cai, Mu and Lee, Yong Jae and others},
  booktitle={ICLR},
  year={2025}
}

@article{han2025multimodal,
  title={Multimodal fusion and vision-language models: A survey for robot vision},
  author={Han, Xiaofeng and Chen, Shunpeng and Fu, Zenghuang and Feng, Zhe and Fan, Lue and An, Dong and Wang, Changwei and Guo, Li and Meng, Weiliang and Zhang, Xiaopeng and others},
  journal={Information Fusion},
  year={2025}
}

@inproceedings{pattnayak2024survey,
  title={Survey of large multimodal model datasets, application categories and taxonomy},
  author={Pattnayak, Priyaranjan and Patel, Hitesh Laxmichand and Kumar, Bhargava and Agarwal, Amit and Banerjee, Ishan and Panda, Srikant and Kumar, Tejaswini},
  booktitle={ICRCV},
  year={2025}
}

@article{wang2025text,
  title={Text-to-cad generation through infusing visual feedback in large language models},
  author={Wang, Ruiyu and Yuan, Yu and Sun, Shizhao and Bian, Jiang},
  journal={ICML},
  year={2025}
}

@misc{ieeexplore,
  author       = {{IEEE}},
  title        = {{IEEE Xplore Digital Library}},
  howpublished = {\url{https://ieeexplore.ieee.org/}},
  year         = {2025},
  note         = {Accessed: May 12th 2025}
}

@article{li2024llava15,
  title={Improved Baselines with Visual Instruction Tuning}, 
  author={Haotian Liu and Chunyuan Li and Yuheng Li and Yong Jae Lee},
  journal={CVPR},
  year={2024}
}

@article{wang2024qwen2vl,
  title={Qwen2-VL: Enhancing Vision-Language Model's Perception of the World at Any Resolution}, 
  author={Peng Wang and Shuai Bai and Sinan Tan and Shijie Wang and Zhihao Fan and Jinze Bai and Keqin Chen and Xuejing Liu and Jialin Wang and Wenbin Ge and Yang Fan and Kai Dang and Mengfei Du and Xuancheng Ren and Rui Men and Dayiheng Liu and Chang Zhou and Jingren Zhou and Junyang Lin},
  year={2024},
  archivePrefix={arXiv},
}

\end{document}